\documentclass[review]{elsarticle}

\usepackage{amssymb}
\usepackage{float}
\usepackage{amsmath}
\usepackage{mathtools}
\usepackage{lmodern}
\usepackage{siunitx}
\usepackage{etoolbox}
\usepackage{lineno}
\usepackage{fixltx2e}
\usepackage{multirow}
\usepackage{multicol}
\usepackage{graphics}
\usepackage{booktabs}
\usepackage{graphicx}

\graphicspath{ {./figures/} }
\usepackage{hyperref}
\usepackage{float}
\usepackage{verbatim} %comments
\usepackage{apalike}
\restylefloat{figure}
\restylefloat{table}

\journal{Expert Systems with Applications}

\biboptions{authoryear}

\begin{document}
\begin{frontmatter}

\begin{titlepage}
\begin{center}
\vspace*{1cm}

\textbf{ \large Improvised Aerial Object Detection approach for YOLOv3 Using Weighted Luminance }

\vspace{1.5cm}

% Author names and affiliations
Sai Ganesh CS$^{a}$ (saiganeshcs@ieee.org), Aouthithiye Barathwaj SR Y$^a$ (aouthithiyebarathwaj@ieee.org),R. Swethaa S$^b$ (swethaa@ieee.org), R. Azhagumurugan$^a$ (azhagumurugan@ieee.org) \\

\hspace{10pt}

\begin{flushleft}
\small  
$^a$ Department of Electrical and Electronics Engineering, Sri Sai Ram Engineering College, Sai Leo Nagar, West Tambaram, Chennai, India \\
$^b$ Department of Electronics and Communication Engineering, Sri Sai Ram Engineering College, Sai Leo Nagar, West Tambaram, Chennai, India \\

\begin{comment}
Clearly indicate who will handle correspondence at all stages of refereeing and publication, also post-publication. Ensure that phone numbers (with country and area code) are provided in addition to the e-mail address and the complete postal address. Contact details must be kept up to date by the corresponding author.
\end{comment}

\vspace{1cm}
\textbf{Corresponding Author:} \\
Sai Ganesh CS \\
Department of Electrical and Electronics Engineering, Sri Sai Ram Engineering College, Sai Leo Nagar, West Tambaram, Chennai, India \\
Tel: +91 8754444820 \\
Email: saiganeshcs@ieee.org

\end{flushleft}        
\end{center}
\end{titlepage}

\title{Improvised Aerial Object Detection approach for YOLOv3 Using Weighted Luminance
}

\author[label1]{Sai Ganesh CS \corref{cor1}}
\ead{bridgesaiganesh@gmail.com}

\author[label1]{Aouthithiye Barathwaj SR Y}
\ead{aouthithiyebarathwaj@gmail.com}

\author[label2]{ Swethaa S }
\ead{swethaa@ieee.org}

\author[label1]{ R. Azhagumurugan }
\ead{azhagumurugan@ieee.org}

\cortext[cor1]{Corresponding author.}
\address[label1]{Department of Electrical and Electronics Engineering, Sri Sai Ram Engineering College, Sai Leo Nagar, West Tambaram, Chennai, India}
\address[label2]{Department of Electronics and Communication Engineering, Sri Sai Ram Engineering College, Sai Leo Nagar, West Tambaram, Chennai, India}

\begin{abstract}
Aerial imaging plays a crucial role in navigation and data acquisition for unmanned aerial vehicles and satellite imaging systems. In recent days, the employment of drones has been escalated in several applications that are not limited to surveillance, delivery systems, aerial warfare, and agricultural activities. Aerial imaging of ground targets is highly challenging because of various factors that affect light propagation through different mediums. Several convolutional neural network-based object detection algorithms that are developed require more robustness when applied in the field of aerial imaging and remote sensing.  In order to handle the adverse effects of light propagation with respect to time and solar radiance, adaptive RGB filters for grayscale imaging based on weighted luminance are introduced that extensively solve the problem of the rayleigh scattering effect. Images of objects that are easily diminished by rayleigh scattering are acquired in various timezones. The acquired images are labelled precisely and subjected to training and validation. The results show that the proposed method detects the object more accurately and efficiently than the traditional YOLOv3 approach.

\end{abstract}

\begin{keyword}
Remote Sensing \sep Computer Vision \sep Convolutional Neural Networks \sep Rayleigh Scattering \sep Grayscale \sep YOLOv3
\end{keyword}

\end{frontmatter}

\section{Introduction}
\label{introduction}

Remote sensing is the process of acquiring data from the earth’s surface without being in actual physical contact with the target. It employs electromagnetic radiation in the form of light to obtain the information about the target on the earth’s surface. It is being spoken of with high values in terms of data acquisition especially where human influence is very low. Remote sensing in the field of defence has been in practice over a long period of time due to its explicitly easier way of communication and data collection over a larger area and we can easily and rapidly process and transmit the data. Remote sensing is considered as the next technological extension to photogrammetry. This enables humans to fetch data beyond normal human eye i.e. infrared, thermal etc. This manages to not only observe objects physically but it also enables them to detect, observe and also record the energy of the target observed without being in a physical contact. Computer vision in the field of remote sensing is gaining a huge popularity as it is used in various applications such as geo-spatial data analysis and it involves using an exponentially large dataset, which is time consuming and involves complex algorithms. The panchromatic camera is the most commonly used device for the purpose of surveying, obtaining images, etc. It is a single lens camera and both CCD (Charge Coupled Devices) arrays and photographic films can be used in the panchromatic camera. It has the ability to sense beyond the visible wavelength and so is named accordingly. When the light is focused on the CCD, it turns the incident photons into voltage and which is then quantized to get the picture in the digital form. The major hurdles to the remote sensing process are the scattering and absorption effects of light which affects the quality of the image during the image acquisition.
\bigbreak

YOLOv3 (You Only Look Once v3) algorithm proposed by Joseph Redmon \cite{r5,y1,y2}, a popular computer vision algorithm is adopted in this article. This algorithm is based on the darknet53 deep learning framework. The darknet53 network comprises of 53 convolutional layers along with batch normalizing layers and Leaky-ReLU activation layers. This network acts as a backbone to the fully-connected YOLOv3 network. Prediction of the coordinates of the bounding boxes is achieved through the method of regression. Dimensional clusters are applied as anchor boxes in order to obtain the required bounding boxes. A special feature of this algorithm is that a single neural network is applied to the full image. Images are divided into multiple regions and coordinates of bounding boxes are obtained. The probabilities for each regions are procured and weighed and the high scoring regions are further processed. In this article, a single channelled YOLOv3, which supports Grayscale images is proposed. This also eliminates several man-made optical illusions, colour illusions and other misinterpretations\cite{2}. 
The bottom surface of the Earth’s atmosphere acts as a reflecting base for the surface and radiations inbound \cite{3,4,5,6,7,8,9,10} This affects the quality of the image that is being collected. Moreover, Earth’s atmosphere consists of various particles such as hydrogen, oxygen, water vapour and many such  particles which attenuates light. In remote sensing, image data acquisition quality is affected by absorption of light from particles such as ozone, water vapour and other atmospheric contents. In addition to the  atmospheric effects, Rayleigh and Mie scattering effects also play a negative role in the field of remote sensing. Shorter wavelength gets scattered more compared to that of larger wavelength. This causes a haze due to the excessively scattered blue light in the captured aerial images hailing to its shorter wavelength. In aerial photography various filters are prevalent to remove these effects but they do not  prove to be of much efficiency and are of higher cost. Also, non-selective scattering of light happens due to clouds and fog, thus making it difficult to interpret the image obtained. These extensive parameters possess an immense challenge to the field of remote sensing especially in the field of image acquisition.. 

\section{Research Methodology}
\label{title_page}
The sky would be black without the presence of atmospheric particles such as NO\textsubscript{2}, O\textsubscript{2} etc. and more importantly the scattering phenomena. During sunrise and sunset, sunlight travels a comparatively longer distance in order to reach the Earth’s atmosphere which tends to give a reddish hue to the color of the sky since shorter wavelengths gets scattered easily \cite{11}. The images acquired at a higher altitude experience Rayleigh’s scattering effect i.e. the shorter wavelengths appear pre-dominant over the others. Thus, the targeted objects in the image dataset of RGB tend to get affected by these factors. Also, some of the images tend to get camouflaged with the surroundings making it completely diminished in size, which pose a serious threat to the areas which border countries and other similar hotspots. Fig. 1 represents the pictorial view of the Rayleigh scattering effect. 
\bigbreak
Fig. 1 (a) This represent the excessive presence of blue wavelength during daytime and Fig. 1(b) represents the sky at sunset and sunrise where red and yellow wavelengths presence largely.
Image pre-processing in this article varies with image acquisition time. This means that Black and white with red filters is used to compensate for the pre-dominant blue color during the day time. Black and white with blue filters are used during the time of sunrise and sunset. Default grayscale images are used during the night time. Thus, all the acquired images are pre-processed prior to the training using the above-mentioned filters. Since the pre-processed images are produced in grayscale, the configuration of the YOLOv3 network is also modified for the suitability of the training. Hence the number of channels in the configuration of YOLOv3 is set to one. The default computing channel in the python interpreter is the BGR (Blue Green Red) channel. So, in this case, the three-channeled BGR images are converted to single-channeled grayscale images which are pre-programmed in the YOLOv3 supporting files \cite{13,14,15,16,17,18,19,20}. 

\begin{figure}[H]
\centering\includegraphics[width=1.0\linewidth]{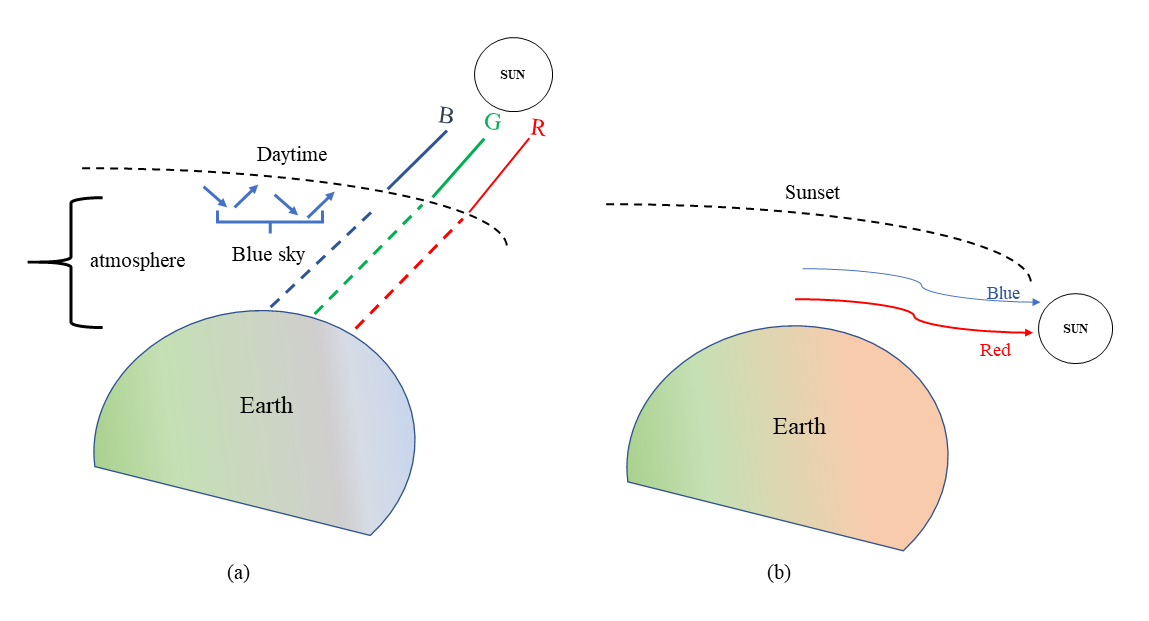}
\caption{Pictorial representation of Rayleigh scattering effect, 1 (a) describes the scattering of blue light in the Earth’s atmosphere, which makes sky appear blue in color during the day time. 1 (b) describes the sunlight traveling a comparatively longer distance to reach the Earth’s surface, making the longer wavelength lights to dominate.}
\end{figure} 

\begin{equation}
\label{eq:emc}
Grayscale(x,y)= 0.3\times R(x,y)+0.1\times G(x,y)+0.5\times B(x,y)
\end{equation}
The traditional conversion of RGB (Red, Green, Blue) channel to grayscale is defined in the Eq (1).

During the day-time, blue light is pre-dominant making the longer wavelengths of light less visible. In order to solve this, grayscale based red filter is adopted to compensate the attenuation of longer wavelengths \cite{g1,g2,g3,g4}. Thus, increase in the intensity of red light is essential. Since green light is not largely affected by the Rayleigh effect, intensity of green is reduced to compensate the attenuation of red light.  The conversion of RGB channeled image to grayscale based red filter is defined in Eq (2).

\begin{equation}
\label{eq:emc}
Grayscale_{weighted} (x,y)= W_r\times R(x,y)+W_g\times G(x,y)+W_b\times B(x,y)
\end{equation}

The weighing factors for each channel is determined individually as given in the equation (2). Where  W\textsubscript{r}, W\textsubscript{g}, W\textsubscript{b}  are the weighing factors of red, green, blue channels respectively. The summation of the above three weighing factors is unity. The RGB histogram of the image is hence  obtained from  which the mean and standard deviation is determined.  The total number of image pixel as percentage of it's frequency is defined as the \textit{Perc} \cite{princi.}. The \textit{Perc} is calculated with respect to the Table 1.

\begin{equation}
\begin{split}
\label{eq:emc}
w_{b}= Perc\times mean           \\
w_{g}=(1-W_b) - (mean + StdDev) \\
w_{r}=1-(w_b+w_g) 
\end{split}
\end{equation}

During the sunset or sunrise, the attenuation of shorter wavelengths occurs, resulting in a reddish tint in the acquired images. Hence, the grayscale based blue filter is adopted in this scenario. Here the intensity of blue is increased and the intensity of red is decreased. The conversion of RGB channeled image into grayscale based blue filter is defined in equation (3). The histogram of sunset/sunrise acquisition image is given in Figure 2 (d) and (g). 
\begin{equation}
\begin{split}
\label{eq:emc}
w_{b}= Perc\times mean \times (2\times stdDev)          \\
w_{g}=(1-W_b) - (avg) \\
w_{r}=1-(w_b+w_g) 
\end{split}
\end{equation}
The conversion of RGB channeled image into grayscale based red filter is defined in equation (4). The R (x, y) is the pixel value corresponding to Red, G (x, y) is the pixel value corresponding to Green and B (x, y) is the pixel value pertaining to Blue, where x and y are pixel locations. The histograms of the original image and the proposed filters are given in Fig. 2.

\begin{table}[H]
\centering
\begin{tabular}{l l l l l}
\hline
\textbf{DN} & \textbf{Npix} & \textbf{Perc}& \textbf{CumNpix} & \textbf{CumPerc}\\
\hline 
0   & 0    & 0.00 & 0     & 0.00 \\
13  & 0    & 0.00 & 0     & 0.00\\
14  & 1    & 0.00 & 1     & 0.00\\
15  & 3    & 0.00 & 4     & 0.01\\
16  & 2    & 0.00 & 6     & 0.01\\
51  & 55   & 0.08 & 627   & 0.86 \\
52  & 59   & 0.08 & 686   & 0.94\\
53  & 94   & 0.13 & 780   & 1.07\\
54  & 138  & 0.19 & 918   & 1.26\\
102 & 1392 & 1.90 & 25118 & 34.36\\
103 & 1719 & 2.35 & 26837 & 36.71 \\
104 & 1162 & 1.59 & 27999 & 38.30\\
105 & 1332 & 1.82 & 29331 & 40.12\\
106 & 1491 & 2.04 & 30822 & 42.16\\
107 & 1685 & 2.31 & 32507 & 44.47\\
108 & 1399 & 1.91 & 33906 & 46.38\\
109 & 1199 & 1.64 & 35105 & 48.02 \\
110 & 1488 & 2.04 & 36593 & 50.06\\
111 & 1460 & 2.00 & 38053 & 52.06\\
163 & 720  & 0.98 & 71461 & 97.76\\
164 & 597  & 0.82 & 72058 & 98.57\\
165 & 416  & 0.57 & 72474 & 99.14 \\
166 & 274  & 0.37 & 72748 & 99.52\\
173 & 3    & 0.00 & 73100 & 100.00\\
174 & 0    & 0.00 & 73100 & 100.00\\
255 & 0    & 0.00 & 73100 & 100.00\\
\hline
\end{tabular}
\caption{Histogram Table}
\end{table}

The model starts training after the preprocessing of the acquired images. Fig. 2. represents the block diagram of the proposed methodology. The image dataset is obtained from LANDSAT 8 via google earth engine which consists of numerous images pertaining to 5 different classes namely, armed vehicles, aircraft, helicopters, tents, and ships/boats. The LANDSAT 8 orbits around the earth at an altitude of 705km. It is inclined at an angle of 98.2 degrees and it takes around 99 minutes to complete one revolution. The images are captured at an altitude of about 20-30 meters above ground level. With the help of LABELIMG the images are annotated in the YOLO format. Thus, supervised learning method is employed. The bounding boxes are drawn along with their respective classes. The training and testing dataset are split as 80\% and 20\% respectively. 
\begin{table}[H]
\centering
\begin{tabular}{l l l l l l}
\hline
\textbf{Dataset} & \textbf{Aircraft} & \textbf{Helicopters}& \textbf{Trucks} & \textbf{Ships} & \textbf{Tents}\\
\hline
Training set   & 220    & 220 & 220     & 220 & 220 \\
Test set    & 25 & 25     & 25 & 25 & 25\\

\hline
\end{tabular}
\caption{ Dataset size and classification}
\end{table}

\begin{table}[H]
\centering
\begin{tabular}{c@{\qquad}ccc@{\qquad}ccc}

  \toprule
  \multirow{2}{*}{\raisebox{-\heavyrulewidth}{\textbf{Image}}} & \multicolumn{2}{c}{\textbf{Original Image}} & \multicolumn{2}{c}{\textbf{Processed Image}} \\
  \cmidrule{2-5}
  &  \textbf{Mean} & \textbf{Std. Deviation} & \textbf{Mean} & \textbf{Std. Deviation}  \\
  \midrule
  $a$ & 128.73 & 50.93  & 98.44 & 51.54\\
  $b$ & 116.92 & 57.98  & 87.60 & 55.57  \\
  $c$ & 102.46 & 56.06  & 75.04 & 52.28\\
  $d$ & 87.76 & 31.92  & 92.38 & 32.56  \\
  $e$ & 126.58 & 28.91  & 139.36 & 26.87\\
  $f$ & 94.01 & 35.61  & 97.86 & 36.42  \\
  \bottomrule
\end{tabular}
\caption{ Mean and standard deviation of the test case images for both original and processed images}
\end{table}

A test case of six images is taken and is processed with the proposed algorithms. The images are obtained from the LandViewer database and Google Earth Engine. The images from NIOP satellite are acquired during sunrise / sunset as the angle of sun elevation is less than 10 degrees. The images obtained from LANDSAT 8 are acquired with the angle of sun elevation greater than 30 degrees. Hence grayscale blue filter is implemented in the NIOP images and grayscale red filter is implemented in LANDSAT 8 images. The images are given in the Figure 2 and the distribution of mean and standard deviation of the histogram of RGB channel is given in Table 3.

\begin{figure}[H]
\centering\includegraphics[width=1.0\linewidth]{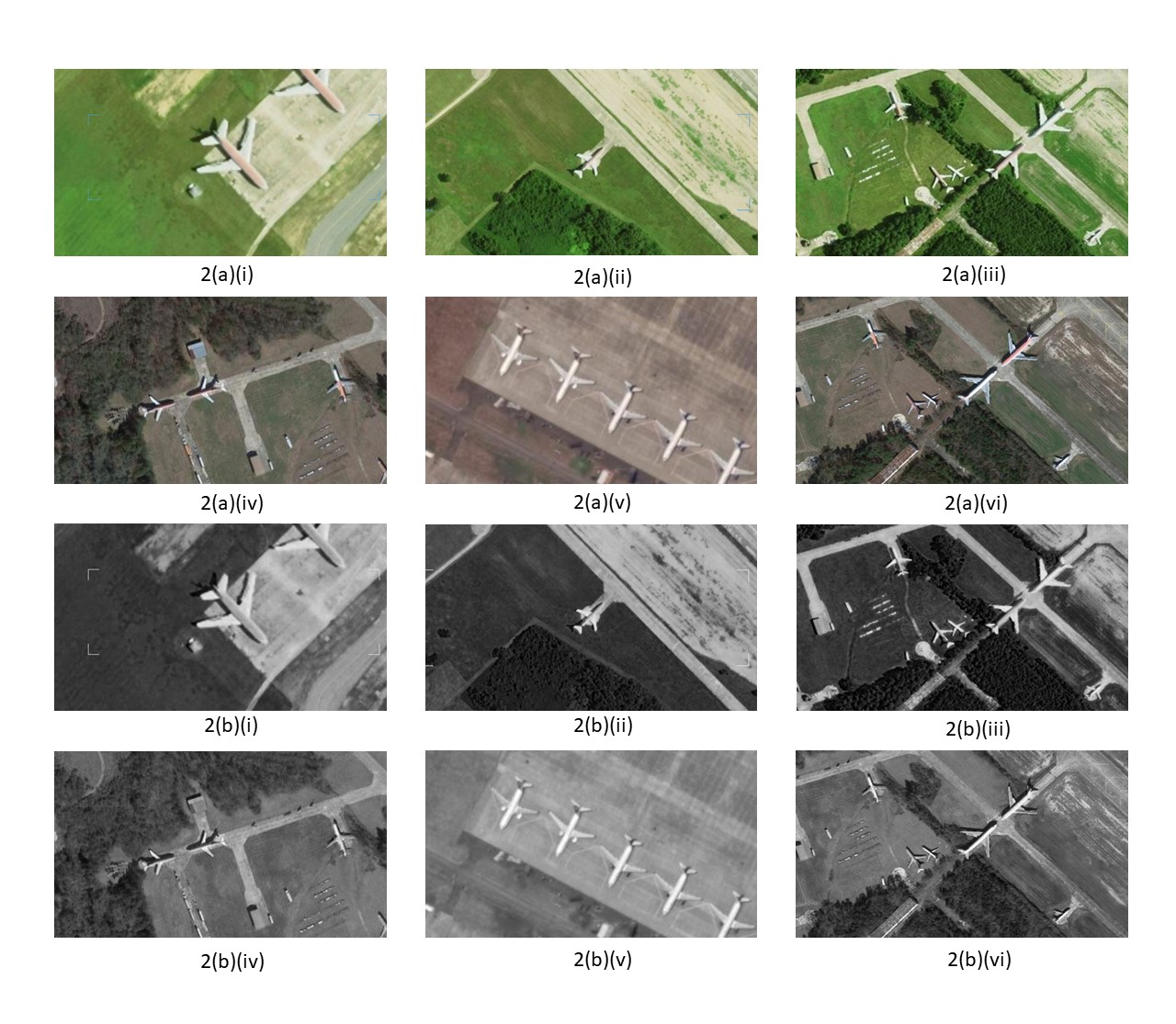}
\caption{Images 2(a)(i), 2(a)(ii) and 2(a)(iii) are the orginal images obtained from NIOP satellite image database. Images 2(a)(iv), 2(a)(v) and 2(a)(vi) are the orginal images obtained from LANDSAT 8 satellite image database 2(b)(i-vi) images are the processed images with the proposed algorithm.}
\end{figure}

\section{Experiment and Analysis}
\label{S:3}
\subsection{Training}

The model is trained in the Google Colaboratory environment using Nvidia P4 16GB GPU, with Core i5-8259U 2.30GHz Quad core CPU with a memory of 8 GB. The learning rate is set to 0.001.The momentum is set to 0.9, with a maximum batch size of 2500. For every single training step, the batch size is set to 64 and 16 subdivisions i.e. 64 images for each step. The training steps were of batch size 2000, 2250 is set at 80\% and 90\%.
\subsection{Results}
The Mean average precision (mAP) and F1-Score are calculated after every 100 iterations. The average loss is calculated after each iteration. The experiment results show that the proposed method is more robust than the YOLOv3 particularly in the field of remote sensing. The mAP results of YOLOv3 started at 67\% and gradually increased to 75.5\% (77\% peak value) over the iterations. In the proposed method, the mAP results started at 30\%, and gradually increased to 84.5\%. The proposed model yields a maximum of 88\% of mAP result. Fig. 3 provides a graph to prove that the proposed method yields better loss and mAP than the traditional YOLOv3.
\begin{figure}
\centering\includegraphics[width=1.0\linewidth]{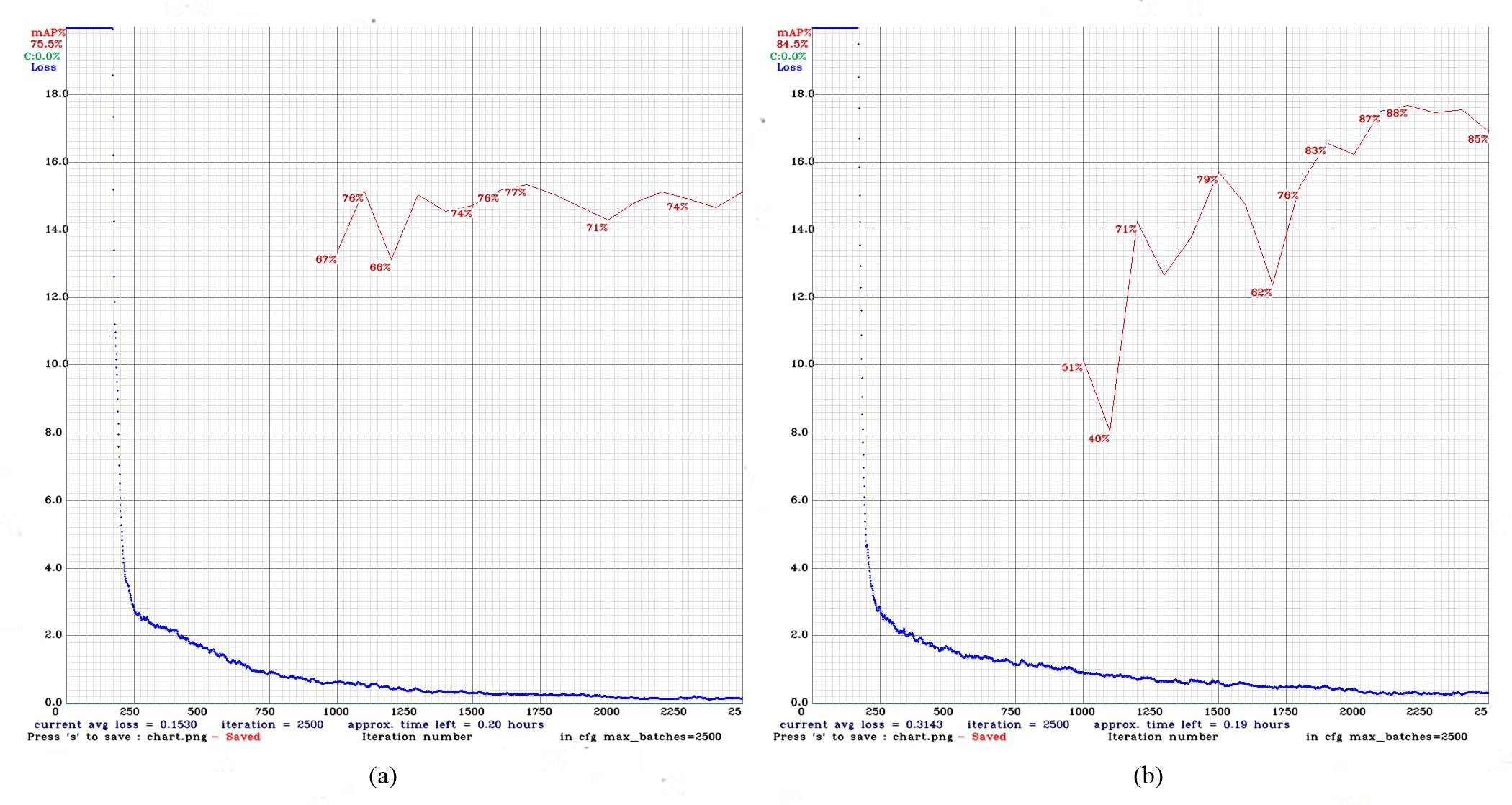}
\caption{. Graphical representation of the mAP and Loss calculations. X-axis represents the number of iterations and Y-axis represents the values for loss and mAP curves. The blue line represents the loss curve and the red line represents mAP. Fig 3 (a) represents the graph for training executed using YOLOv3 and Fig 3 (b) represents the graph of the training executed using the proposed method}
\end{figure}

\begin{table}[H]
\centering
\begin{tabular}{l l l l}
\hline
\textbf{Method} & \textbf{Minimum Loss} & \textbf{mAP}& \textbf{Peak mAP}\\
\hline
YOLOv3   & 0.153    & 75.5\% & 77\%  \\
Proposed Method    & 0.314 & 84.5\%     & 88\% \\

\hline
\end{tabular}
\caption{Experiment results of the implementation of YOLOv3 and the proposed methodologies computing the minimum training loss and Mean Average Precision }
\end{table}

\subsection{Deployment}
The deployment of the model is executed in PyCharm IDE with the trained configuration and weights. With the help of time module library from Python, we apply the respective filters corresponding to its time of image acquisition. The confidence score for the object detection is set to 0.30. Fig. 4 compares the images predicted by the YOLOv3 based approach and the proposed approach. 

\begin{figure}[H]
\centering\includegraphics[width=1.0\linewidth]{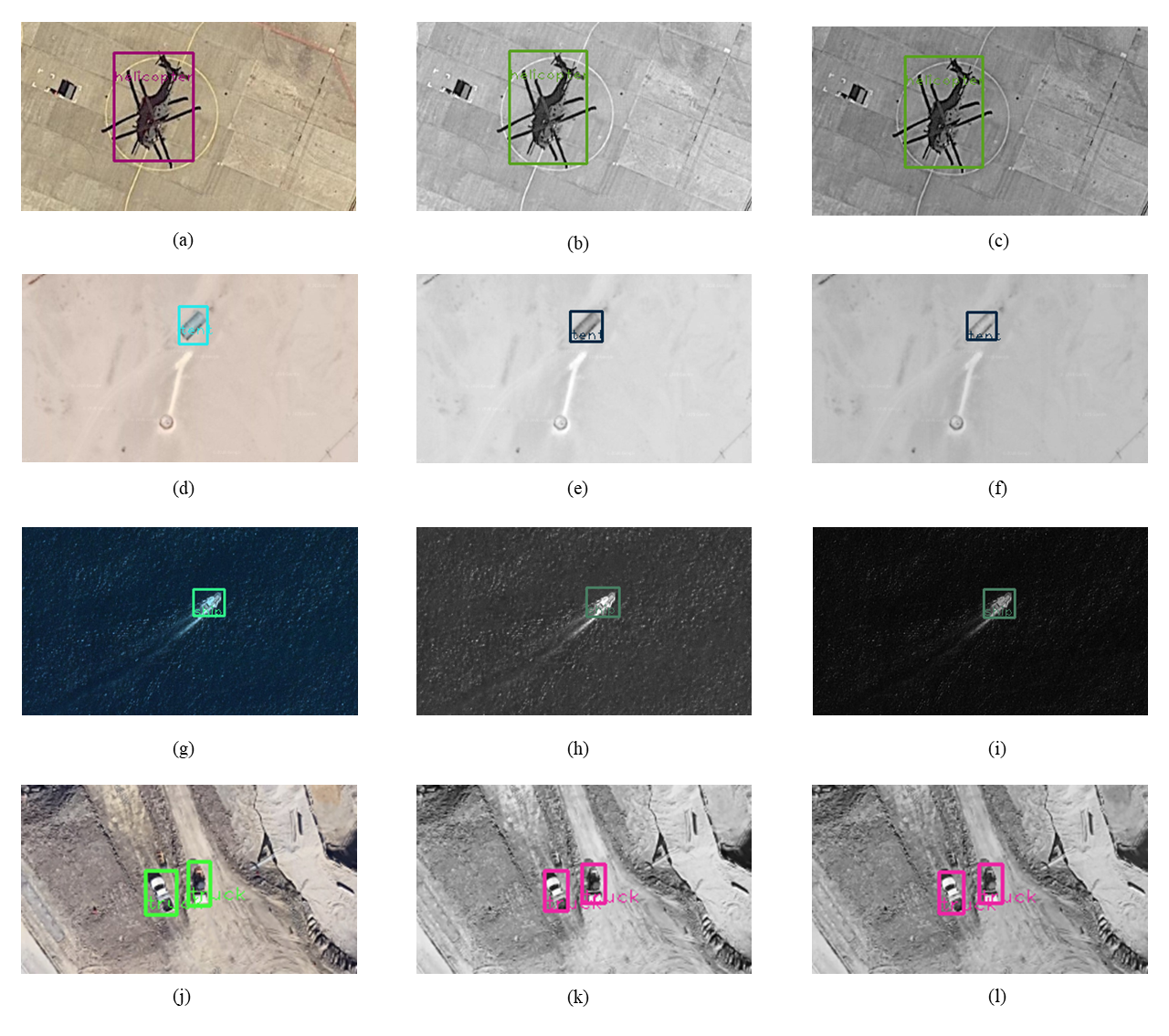}
\caption{List of Deployment images with different filters. 4 (a), (d), (g), (j) are original acquired images. 4 (b), (e), (h), (k) are images with grayscale based red filter. 4 (c), (f), (i), (l) are images with grayscale based blue filter.}
\end{figure}

\section{Conclusion}
\label{S:4}
In this article, grayscale based algorithms are specifically implemented with respect to the sun's angle of elevation. The images are obtained from NIOP and LANDSAT 8 satellite imagery using LandViewer database. All the acquired images are analyzed and implemented with respect to its adaptable algorithm. After the preprocessing of images, the images are trained with deep learning based YOLOv3 algorithm for target detection. The training results are discussed and the model is successfully deployed using PyCharm IDE. the future scope of this project aims to program a machine that alarms any trespasses detection and sends a warning through email and mobile applications.

\section*{Acknowledgements}
The authors would like to thank the management of Sri Sai Ram Engineering College who gave technical support to this research work. The authors would like to show their gratitude towards Dr. T. Porselvi and Dr. B. Meenakshi for their continuous support and guidance.

\bibliography{sample}
\end{document}